\documentclass{article}

\usepackage{arxiv}

\usepackage{times}  

\usepackage{booktabs} 

\usepackage{epsfig}
\usepackage{graphicx}
\usepackage{amsmath}
\usepackage{amssymb}
\usepackage{subfigure}
\usepackage{algorithm}
\usepackage{algorithmic}
\usepackage{natbib}

\newcommand{\xs}{x^s}
\newcommand{\xt}{x^t}
\newcommand{\ys}{y^s}
\newcommand{\yt}{y^t}

\newcommand{\xucomment}[1]{}

\usepackage[utf8]{inputenc} 
\usepackage[T1]{fontenc}    
\usepackage{hyperref}       
\usepackage{url}            
\usepackage{booktabs}       
\usepackage{amsfonts}       
\usepackage{nicefrac}       
\usepackage{microtype}      
\usepackage{lipsum}

\title{DART: Domain-Adversarial Residual-Transfer Networks for Unsupervised Cross-Domain Image Classification}

\author{
  Xianghong Fang \\
  SMILE Lab \\
 Univ. of Electronic Sci \& Tech.of China  \\
  Chengdu, Sichuan 611731 China \\
   \AND 
   Haoli Bai \\
   Dept Computer Science and Engineering  \\ 
   The Chinese University of Hong Kong  \\ 
   Shatin, N. T.  Hong Kong,  China  \\
 \AND 
 Ziyi Guo \\
   SMILE Lab, Sch. of Computer Science and Engineering \\ 
 University of Electronic Science and Technology of China  \\
  Chengdu, Sichuan 611731 China \\
 \AND
  Bin Shen \\
Google Inc. \\
\AND 
Steven Hoi \\
School of Information Systems (SIS) \\
  Singapore Management University \\
\AND 
Zenglin Xu\thanks{Corresponding author.}  \\
  SMILE Lab, Sch. of Computer Science and Engineering \\ 
 University of Electronic Science and Technology of China  \\
  Chengdu, Sichuan 611731 China \\
  \texttt{zenglin@gmail.com} 
}

\begin{document}
\maketitle

\begin{abstract}
The accuracy of deep learning (e.g., convolutional neural networks) for an image classification task critically relies on the amount of labeled training data. Aiming to solve an image classification task on a new domain that lacks labeled data but gains access to cheaply available unlabeled data, unsupervised domain adaptation is a promising technique to boost the performance without incurring extra labeling cost, by assuming images from different domains share some invariant characteristics. In this paper, we propose a new unsupervised domain adaptation method named Domain-Adversarial Residual-Transfer (DART) learning of Deep Neural Networks to tackle cross-domain image classification tasks. In contrast to the existing unsupervised domain adaption approaches, the proposed DART not only learns domain-invariant features via adversarial training, but also achieves robust domain-adaptive classification via a residual-transfer strategy, all in an end-to-end training framework. We evaluate the performance of the proposed method for cross-domain image classification tasks on several well-known benchmark data sets, in which our method clearly outperforms the state-of-the-art approaches.

\end{abstract}


\section{Introduction}

Recent years have witnessed remarkable successes of deep learning methods, especially the Deep Convolutional Neural Networks (CNN), for various image classification and visual recognition tasks in multimedia and computer vision domains. The successes of deep neural networks for image classification tasks critically rely on large amounts of labeled data, usually leading to networks with millions of parameters~\citep{Lin2014MicrosoftCC,Russakovsky2015ImageNetLS}. In practice, when solving an image categorization task in a new domain, collecting a large amount of labeled data is often difficult or very expensive, yet a large amount of unlabeled data is cheaply available.
How to leverage the rich amount of unlabeled data in the target domain and to resort to an existing classification task from a source domain has become an important research topic, which is often known as  unsupervised domain adaption or transfer learning~\citep{Pan2010ASO,PanYLM17,Wang2018DeepVD}. The goal of this work is to explore new unsupervised domain adaption techniques for cross-domain image classification tasks. 

Unsupervised domain adaptation has been actively studied in literature. One of the dominating approaches seeks to bridge source domain and target domain through learning a domain-invariant representation, and then training an adaptive classifier on the target domain by exploiting knowledge from the source domain. Following such kind of principle, several previous works have proposed to learn transferable features with deep neural networks~\citep{Long2015LearningTF,Tzeng2014DeepDC,Long2016UnsupervisedDA,Long2017DeepTL,Zhuo:2017:DUC}, by minimizing a distance metric of domain discrepancy, such as  Maximum Mean Discrepancy (MMD)~\citep{Gretton2006AKM}. Recently,   inspired by Generative Adversarial Networks (GANs)~\citep{Goodfellow2014GenerativeAN}, a surge of emerging studies proposed to apply adversarial learning for unsupervised domain adaptation~\citep{Tzeng2017AdversarialDD,Ganin2015UnsupervisedDA,Liu2016CoupledGA,Liu2017UnsupervisedIT,Taigman2016UnsupervisedCI,Bousmalis2016DomainSN,ganin2016domain}, validating the advantages of adversarial learning over traditional approaches in minimizing domain discrepancy and obtained new state-of-the-art results on benchmark datasets for unsupervised domain adaptation. 

Among the emerging GAN-inspired approaches, DANN~\citep{ganin2016domain} represents an important milestone. Based on the common low-dimensional features shared by both source and target domains, DANN introduces a domain classifier borrowing the idea from GAN to help learn transferable features. The domain classifier and the feature representation learner are trained adversarially, where the former strives to discriminate the source domain from the target domain, while the latter tries to learn domain indistinguishable features from both domains. Then a label classifier is deployed to predict the labels of samples from both domains with the learned domain-invariant features.

\begin{figure}[h]
\centering
\includegraphics[width=7cm]{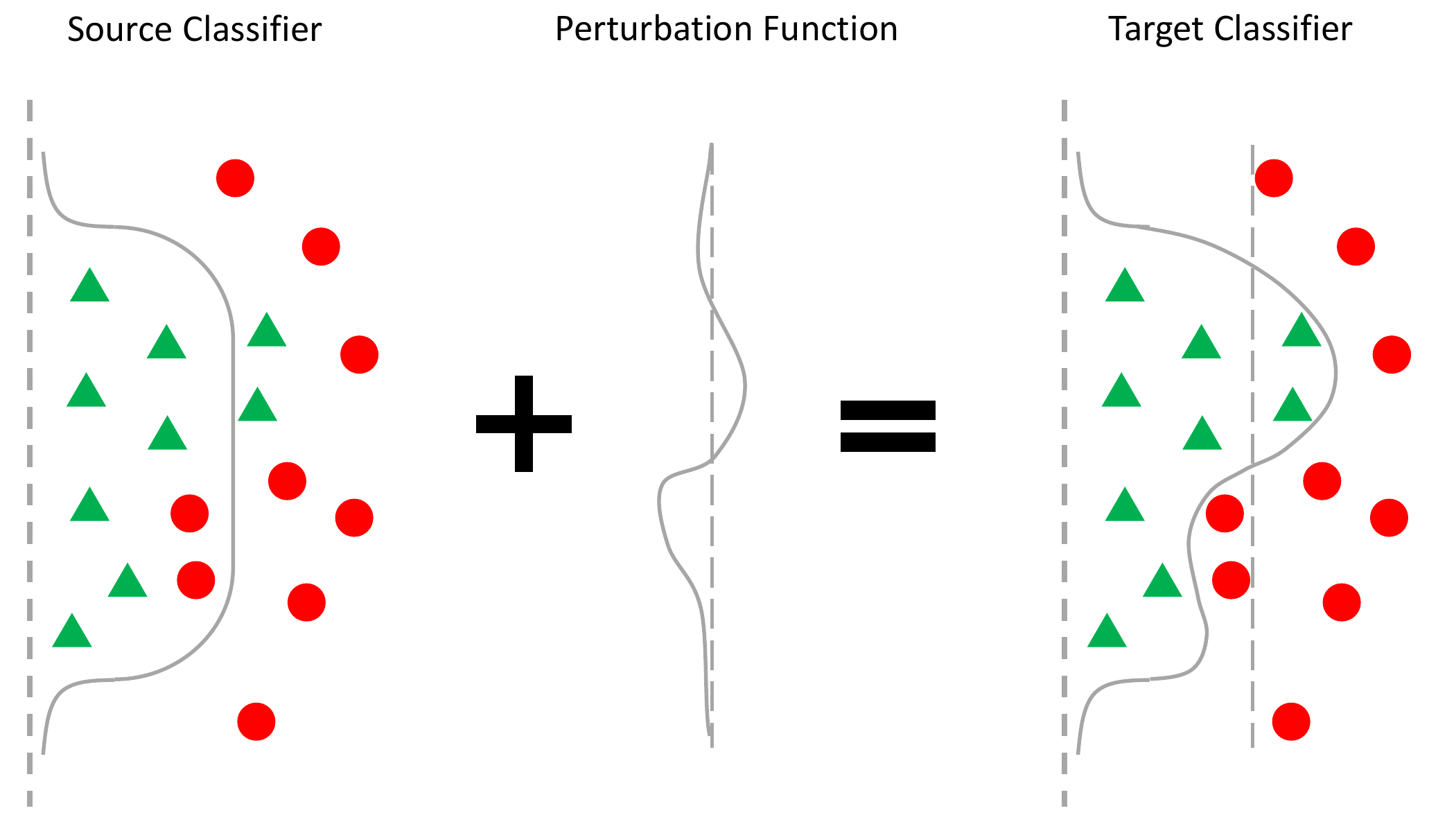}
\caption{A well-trained source classifier may fail to classify images in the target domain correctly. By adding a perturbation function, the target classifier corrects the mistakes made by the source classifier on the target domain. Here the red dots and green triangles denote image samples from the target domain. }
\label{source classifier and target classifier}
\end{figure}

Despite the success of DANN, it has two major limitations. First, it assumes that the image class label classifier of the source domain can be directly applied to the target domain. However, in practice there could be some small shifts across the label classifiers in two domains, since the image classification tasks on both two domains can be quite different. 
An intuitive example is illustrated in the first column of Figure~\ref{source classifier and target classifier}, where the source classifier fails to correctly classify data from the target domain.
Second, the label information of labeled training data in the source domain is not exploited when learning the domain-invariant features. In other words, minimizing the discrepancy of marginal distributions (i.e., without exploiting label information) may only lead to some restrictive representations lacking strong class discriminative ability.

In this paper, we propose a new unsupervised domain adaptation approach named Domain-Adversarial Residual-Transfer (DART) learning for training Deep Neural Networks to tackle cross-domain image classification tasks. Specifically, the proposed DART architecture consists of three key components: image feature extractors, image  label classifiers, and a domain classifier, which will be discussed in detail in Section~\ref{sec:model}. DART inherits all the advantages of DANN-based network architectures for domain-adversarial learning, but makes the following two important improvements. 

First, in order to model the shifts across the label classifiers in different domains, we introduce a perturbation function across the label classifiers, and insert the ResNet~\citep{He2016DeepRL} into the source label classifier to learn the perturbation function, since ResNet has demonstrated superior advantages in modelling the perturbation via a shortcut connection as shown in~\citep{Long2016UnsupervisedDA}. With the learned perturbation function, the label classifier could be more robust and accurate. An intuitive illustration is shown in Figure~\ref{source classifier and target classifier}, where the target classifier with the perturbation function correctly classifies image samples from the target domain.

Second, to learn representations that are more discriminative and robust, we exploit the joint distributions of image features and class labels to align the source domain and the target domain. Specifically, our model is established based on a relaxed but more general assumption in that the joint distribution of both image data and class labels in the source domain is different from that joint distribution in the target domain. Therefore, our model seeks to reduce their joint discrepancy when learning the common feature space. Notice that class labels in the target domain are replaced by pseudo labels predicted by existing classifiers since these labels are unavailable during training in unsupervised domain adaption. Instead of minimizing the discrepancy on marginal distributions, minimizing the joint discrepancy learns more discriminative domain-invariant features, as the joint distributions leverage the additional label information. We further regularize the model by minimizing the entropy of the predicted labels of the target domain, which ensures classification predictions from the target classifier stay away from low-density regions. 

As a summary, this work makes the following major contributions:
\begin{enumerate}
\item Our work incorporates the learning of the perturbation function across label classifiers into the adversarial transfer network. This makes the transfer learning more adaptable to real-world image domain adaptation tasks. 
\item Our work focuses on minimizing the discrepancy of the joint distributions of both image samples and class labels in the adversarial learning scheme, thus is able to yield domain-invariant features which are more discriminative . 

\item We conduct extensive evaluation of cross-domain image classification on several benchmarks, in which DART clearly outperforms the state-of-the-art methods on most cases. 
\end{enumerate}

\section{Cross-Domain Image Classification} \label{sec:model}

\subsection{Problem Setting} 

\begin{table}[h]
	\centering
	\caption{Notations}
	\begin{tabular}{l|c}
            \hline
			$\xs_i,\xt_i$ & image samples from source \& target domain\\\hline
			$\ys_i, \yt_i$ & labels from source \& target domain\\\hline
			$\mathcal{D}_s,\mathcal{D}_t$ & Data of source and target domain \\\hline
			$d^s_i, d^t_i$ & domain labels for source \& target domain \\\hline
			$G_f(\cdot), G_d(\cdot)$ & feature extractor \& domain classifier\\\hline
			$ \theta $ & parameters for certain function \\\hline
			$\sigma(\cdot)$ & softmax function
			\\\hline
	\end{tabular}
	\label{notations}
\end{table}

We consider a cross-domain image classification task by following a common setting of unsupervised domain adaption. Specifically, consider a collection of ${N_s}$ image samples and class labels from a source domain $\mathcal{D}_s = \{(\xs_i,\ys_i)\}_{i=1}^{N_s}$, and ${N_t}$ unlabeled image samples from a target domain $\mathcal{D}_t =\{\xt_i\}_{i=1}^{N_t}$. The goal of unsupervised cross-domain image classification is to adapt a classifier trained using only labeled data from the source domain and unlabeled data from the target domain, such that the domain-adaptive classifier can correctly predict the labels $y^t$ of images from the target domain. 

It is important to note that a major challenge of this problem is the absence of labeled data in the target domain, making it different and much more challenging from many existing supervised transfer learning problems.
Another key challenge of unsupervised domain adaption is that the source image classifier trained on the source domain $\mathcal{D}_s$ cannot be directly applied to solve the image classification tasks in the target domain $\mathcal{D}_t$, because the image data between the source domain and the target domain can have large discrepancy, and their joint and marginal distributions are different, i.e. $p(\xt, \yt)\neq p(\xs, \ys)$ and $p(\xt)\neq p(\xs)$, where $\yt$ is the true underlying target class labels.

\subsection{Overview of Proposed DART} 

To minimize the discrepancy of the source domain and the target domain effectively, we propose the Domain-Adversarial Residual-Transfer learning (DART) of training Deep Neural Networks for unsupervised domain adaptation, as shown in Figure~2. The proposed DART method is based on two assumptions. 

First of all, DART assumes the joint distributions of labels and high-level features of data should be similar. The high-level features are extracted by a feature extractor $G_f(\cdot)$ parameterized by $\theta_f$. Then a Kronecker product is applied on high-level features and the label information to obtain joint representations. Finally, these joint representations are collectively embedded into a domain classifier $G_d(\cdot)$ parameterized by $\theta_d$ to ensure $p(G_f(\xt), \hat{y}^t) \approx p(G_f(\xs), \ys)$, where $\hat{y}^t$ represents the predicted label.

\begin{figure*}[ht]
\centering
\includegraphics[width=.7\textwidth]{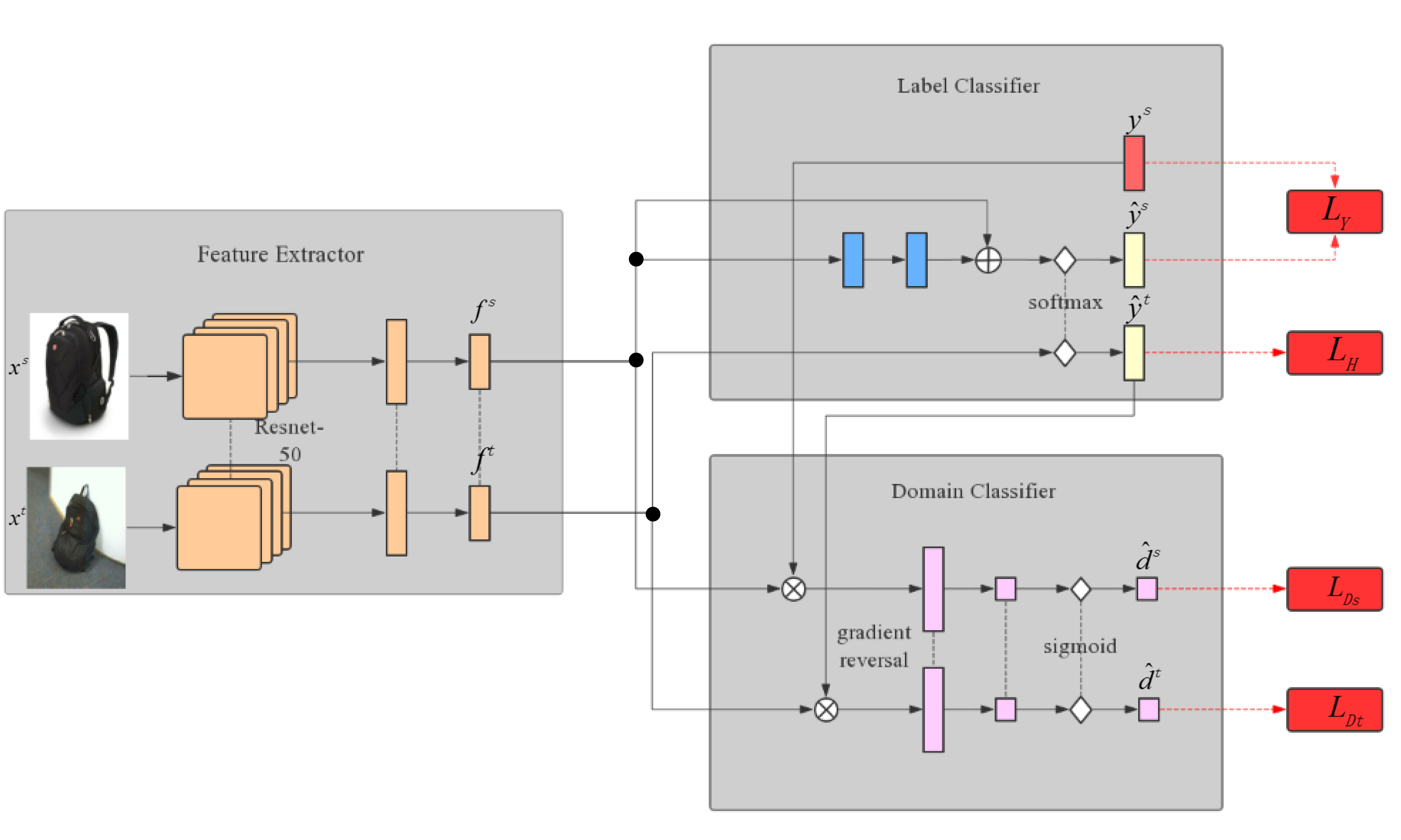}\label{DART-network}
\caption{The architectures of Domain-Adversarial Residual-Transfer Networks (DART), which consists of a deep feature extractor (yellow), a deep label classifier (blue) and a domain classifier (purple). $\oplus$ means the plus operator while $\otimes$ represents the Kronecker product. The crossing lines with black points indicate connections, while crossing lines without black points are independent and not connected. The domain classifier is connected to the feature extractor via a gradient reversal layer. 
During the training process, we minimize the label prediction loss (source examples for $\mathcal{L}_Y$ and target examples for $\mathcal{L}_H$) and the domain classification loss (for all samples).
Gradient reversal ensures that the joint distributions over the two domains are made similar, thus resulting in the domain-invariant features.}
\vspace{-0.3cm}
\end{figure*}


Second, DART assumes the label classifier for the target domain differs from that of the source domain, and the difference between the two classifiers can be modeled by the following $$p(\yt|\xt)=p(\ys|\xs) + \epsilon,$$ where $\epsilon$ is a perturbation function across the source label classifier $G_s(\cdot)$ and the target label classifier $G_t(\cdot)$. In order to learn the perturbation function, one way is to introduce residual layers~\citep{He2016DeepRL} parameterized by $\theta_{r}$ by inserting into the source label classifier. Note that the predicted labels are $\hat{y}_t = G_t(G_f(x_t))$. In the following sections, we introduce each module in detail.






\subsection{Domain Adversarial Training for Joint Distribution}
Minimizing the domain discrepancy is crucial in learning domain-invariant features. A number of previous work seek to  minimize over a metric, i.e., MMD~\citep{Gretton2006AKM}. These methods have been proved experimentally effective, however, they suffer from large amount of hyper-parameters and thereon difficulty in training. A more elegant way is to operate on the architecture of the neural network. DANN~\citep{ganin2016domain} is a representative work in which domain discrepancy is reduced in an adversarial way, leading to an easier and faster training process.

Inspired by DANN, we devise an adversarial transfer network to collectively distinguish the joint distribution $ p(G_f(\xs), \ys)$ and $p(G_f(\xt), \hat{y}^t)$. Specifically, for each domain, we fuse the high-level features from the feature extractor and labels together via a Kronecker product, i.e. $G_f(x_i^s) \otimes y_i^s$ and $G_f(x_i^t)\otimes \hat{y}_i^t$,
and then embed them into the domain classifier $G_d(\cdot)$ so as to minimize the discrepancy of their joint distributions in an adversarial way. The feature extractor seeks to learn indistinguishable features from the source domain and the target domain, while the domain classifier is trained to discriminate the domain of features correctly.
In the domain classifier, we manually label the source domain as $d^s_i = 1$, and the target domain as $d^t_i = 0$.
We introduce the~\textbf{gradient reversal layer} (GRL) as proposed in~\citep{ganin2016domain} for a more feasible adversarial training scheme. Given a hyper-parameter $\lambda$ and a function $f(v)$ and , the GRL can be viewed as the function $g(f(v);\lambda)=f(v)$ with its gradient $\frac{d}{dv}g(f(v);\lambda)=-\lambda \frac{d}{dv}f(v)$. With GRL, we could minimize over $\theta_f$, $\theta_d$ directly by the standard back propagation.
The output $\hat{d}^s_i$ and $\hat{d}^t_i$ of the domain classifier can be written as follows:
\begin{align}
\hat{d}^s_i &= G_d(g(G_f(x_i^s) \otimes y_i^s)), \\ \nonumber
\hat{d}^t_i &= G_d(g(G_f(x_i^t)\otimes \hat{y}_i^t)).
\end{align}

To attain precise domain prediction of joint representations,
we define the loss of domain classifier as follows:
\begin{align}
\mathcal{L}_D &= \mathcal{L}_{Ds}+\mathcal{L}_{Dt} \nonumber\\
    &=-\frac{1}{N_s}\sum_{i=1}^{N_s}d^s_i\log{\hat{d}^s_i}- \frac{1}{N_t}\sum_{i=1}^{N_t}(1-d^t_i)\log{(1-\hat{d}^t_i)}.
\end{align}

\subsection{Residual Transfer Learning for Label Classifier  Perturbation}
A key point in transfer learning is to predict labels in the target domain based on the domain-invariant features. A common assumption for label classifiers is that given the domain-invariant features, the conditional distribution of labels are the equal, i.e., $p(\yt|\xt)=p(\ys|\xs)$. However, this might be insufficient to capture the underlying  perturbation of label classifiers across different domains. Hence, we assume $p(\yt|\xt)=p(\ys|\xs) + \epsilon$, and consider the modelling of the  perturbation between label classifiers.
The residual layers~\citep{He2016DeepRL} has shown its superior advantages in modelling the perturbation via a shortcut connection in RTN~\citep{Long2016UnsupervisedDA}, and it can be concluded by
$f_s(x;\theta_r)= f_t(x) +\Delta f(x;\theta_r)$, where $\Delta f(x)$ is the perturbation function parameterized by $\theta_r$ and $f_t(\cdot)$ is the identity function. Residual layers ensure the output to satisfy $|\Delta f(x)|\ll |f_t(x)| \approx |f_s(x)|$, as verified in~\citep{Long2016UnsupervisedDA}.
We set $G_s(G_f(\xs)) \triangleq \sigma(f_s(G_f(\xs)))$ and
$G_t(G_f(\xt)) \triangleq \sigma(f_t(G_f(\xt))) = \sigma(G_f(\xt))$,  where $\sigma(\cdot)$ is the softmax function to give specific predicted probabilities.

For the source label classifier, the loss function can be easily computed as
\begin{align}
\mathcal{L}_Y&=-\frac{1}{N_s}\sum_{i=1}^{N_s}\mathcal{L}(G_s(G_f(x^s_i)),y^s_i)\nonumber\\
   &=-\sum_{i=1}^{N_s}\{y^s_i\log{G_s(G_f(x^s_i))}\}.
\end{align}

For the target label classifier, the learned label classifier may fail in fitting the possibilities of ground truth target labels well. To tackle this problem, following~\citep{Grandvalet2004SemisupervisedLB}, we further minimize the entropy of class-conditional distribution
$p(y^t_i=j|x^t_i)$ as
\begin{equation}
\mathcal{L}_H=-\frac{1}{N_t}\sum_{i=1}^{N_t}\sum_{j=1}^{c} p(y^t_i=j|x^t_i) \log{p(y^t_i=j|x^t_i))}.
\end{equation}
where $c$ represents the number of classes, and $p(y^t_i=j|x^t_i)$  can be obtained by $p(y^t_i|x^t_i) = G_t(G_f(x_i^t))$.
By minimizing the entropy penalty, the target classifier $G_t$ would adjust itself to enlarge the difference of possibilities among the predictions, and thereon predict more indicative labels.

Finally, the overall objective function of our model is
\begin{align}\label{total loss}
\mathcal{L} = \mathcal{L}_Y+\alpha \mathcal{L}_H +\beta \mathcal{L}_D.
\end{align}
where $\alpha$, $\beta$ are the trade-off regularizers for the entropy penalty and the domain classification. Our goal is to find the optimal parameters $\theta_f^*$, $\theta_d^*$, $\theta_r^*$ by minimizing Equation \ref{total loss}.

\xucomment{
:
\begin{align}
&\theta_f^*,\theta_r^*, \theta_d^* = \mathop{\arg\min_{\theta_f,\theta_r, \theta_d}} \mathcal{L}.
\end{align}
}

To clearly present our DART model, we represent the pseudo-code in Algorithm \ref{DART algorithm}.




\begin{algorithm}[t] 
\caption{\small The algorithm description for DART.}


\begin{algorithmic}
\STATE Require:
\begin{itemize}
    \item source  samples and labels ($x_i^s,y_i^s$) and target samples $x_i^t$
    \item domain classifier label $d_i^s =1$, $d_i^t =0$
    \item trade-off parameter $\alpha$ and $\beta$ for entropy penalty and domain classification respectively
    and hyper-parameter $\lambda(t)$ for the gradient reversal layer function $g(\cdot;\lambda(t))$.
    \item the feature extractor $G_f(\cdot);\theta_f)$  with parameters $\theta_f$, 
    \item the domain classifier $G_d(\cdot)$ with parameters $\theta_d$
    \item the source classifier $G_s(\cdot;\theta_r)$
    with parameters $\theta_r$ and target classifier
    $G_t(\cdot)$
\end{itemize}
\FOR{$t$ in $[1,training\_steps]$ } 
  \FOR{minibatch $A,B$ } 
  \STATE Extract features from source samples and target samples:
  \STATE $f^s_i =G_f(x^s_{i\in{A}};\theta_f,t), f^t_i = G_f(x^t_{i\in{B}};\theta_f,t)$
  \STATE Obtain source label prediction and target label prediction:
  \STATE$\hat{y}^s_i = G_s(f^s_i;\theta_r,t), \hat{y}^t_i = G_t(f^t_i;t)$
  \STATE Fuse features and labels with a Kronecker product to represent joint representation:
  \STATE$z^s_i = f^s_i \otimes y^s_i, z^t_i = f^t_i \otimes \hat{y}^t_i$
  \STATE Embed joint representation into domain classifier:
  \STATE $d_i^s = G_d(g(z^s_i;\lambda(t));\theta_d,t), d_i^t = G_d(g(z^t_i;\lambda(t));\theta_d,t)$
  \STATE Update loss function of domain classifier:  
  \STATE $\mathcal{L}_D = -\frac{1}{|A|}\sum_{i\in A}d^s_i\log{\hat{d}^s_i}$
  \STATE $\qquad\:\:\:-\frac{1}{|B|}\sum_{i \in B}(1-d^t_i)\log{(1-\hat{d}^t_i)}$
  \STATE Update loss function of source classifier: \STATE$\mathcal{L}_Y =-\frac{1}{|A|}\sum_{i\in A}y_i^s \log{\hat{y}^s_i}$
  \STATE Update loss function of target classifier: \STATE$\mathcal{L}_H =-\frac{1}{|B|}\sum_{i\in B}\hat{y}^t_i \log{\hat{y}^t_i}$
  \STATE Update overall objective loss function: \STATE$\mathcal{L} = \mathcal{L}_Y+\alpha \mathcal{L}_H +\beta \mathcal{L}_D$
  \STATE  Update $\theta_f$,$\theta_d$ and $\theta_r$ using SGD optimizer
  \ENDFOR
\ENDFOR \newline
\textbf{Return} $\hat{\theta}_f,\theta_d,\theta_r$
\end{algorithmic}
\label{DART algorithm}
\end{algorithm}


\section{Experiment} \label{sec:exp}

We evaluate the proposed DART against several state-of-the-art baselines on unsupervised domain adaptation problems. Codes and datasets will be released.

\subsection{Datasets and Baselines}
$\textbf{\emph{USPS}}  \leftrightarrow  \textbf{\emph{MNIST}}:$ MNIST~\citep{lecun1998gradient}
contains 60000 training digit images and 10000 test digit images, and USPS~\citep{Denker1988NeuralNR} contains 7291 training images and 2007 test images. 
For the transfer task from MNIST to USPS, we use the labeled MNIST dataset as the source domain and use unlabeled USPS dataset as the target domain, and vice versa for the transfer task from USPS to MNIST.




\textbf{\emph{Office-31}}\footnote{http://office31.com.my}~\citep{Saenko2010AdaptingVC} is a standard benchmark for unsupervised domain adaptation. It consists of 4652 images and 31 common categories collected from three different domains: \emph{Amazon} (\textbf{A}) which contains 2817 images from amazon.com, \emph{DSLR} (\textbf{D}) which contains 498 images from digital SLR camera and \emph{Webcam} (\textbf{W}) which contains 795 images from the web camera.
We evaluate all methods on the following six transfer tasks: $\textbf{A} \to \textbf{W}$, $\textbf{D} \to \textbf{W}$, $\textbf{W} \to \textbf{D}$, $\textbf{A} \to \textbf{D}$, $\textbf{D} \to \textbf{A}$, $\textbf{W} \to \textbf{A}$, as done in~\citep{Long2017DeepTL,Long2016UnsupervisedDA}. 

\textbf{\emph{ImageCLEF-DA}}\footnote{http://imageclef.org/2014/adaptation} is a benchmark dataset for the \emph{ImageCLEF 2014} domain adaptation challenge, consisting of three public domains: \emph{Caltech-256}(\textbf{C}), \emph{ImageNet ILSVRC 2012}(\textbf{I}), and \emph{ Pascal VOC 2012}(\textbf{P}). Each domain contains 12 categories and  each category has 50 images. 
We also consider all the possible six transfer tasks: $\textbf{I} \to \textbf{P}$, $\textbf{P} \to \textbf{I}$, $\textbf{I} \to \textbf{C}$, $\textbf{C} \to \textbf{I}$, $\textbf{C} \to \textbf{P}$, $\textbf{P} \to \textbf{C}$, as done in~\citep{Long2017DeepTL}.

For MNIST to USPS and USPS to MNIST, we compare with three recent unsupervised domain adaptation algorithms: CoGAN~\citep{Liu2016CoupledGA}, pixelDA~\citep{Bousmalis2016UPD}, and UNIT~\citep{Liu2017UnsupervisedIT}.
CoGAN and UNIT seek to learn the indistinguishable features from the discriminator.
PixelDA is an effective method in unsupervised domain adaptation in which a generator is used to map data from the source domain to the target domain.
We choose these GANs-based baselines since it has been illustrated in~\citep{Goodfellow2014GenerativeAN} that GANs have more advantages over conventional kernel method (e.g., MMD) on reducing distribution discrepancy of domains.


For \emph{Office-31} and \emph{ImageCLEF-DA} datasets, in order to have a fair comparison with the latest algorithms in unsupervised domain adaptatoin, we choose the same baselines as reported in Joint Adaptation Network (JAN)~\citep{Long2017DeepTL}. Aside from JAN, other baselines include Transfer Component Analysis (TCA)~\citep{Pan2009DomainAV},
Geodesic Flow Kernel (GFK)~\citep{Gong2012GeodesicFK}, 
ResNet~\citep{He2016DeepRL},
Deep Domain Confusion (DDC)~\citep{Tzeng2014DeepDC}, 
Deep Adaptation Network (DAN)~\citep{Long2015LearningTF},
Residual Transfer Network (RTN)~\citep{Long2016UnsupervisedDA},
Domain-Adversarial Training of Neural Networks (DANN)~\citep{Ganin2015UnsupervisedDA}.



\begin{figure}[h]
\centering
\subfigure[USPS and MNIST samples]{
\includegraphics[width=0.45\textwidth]{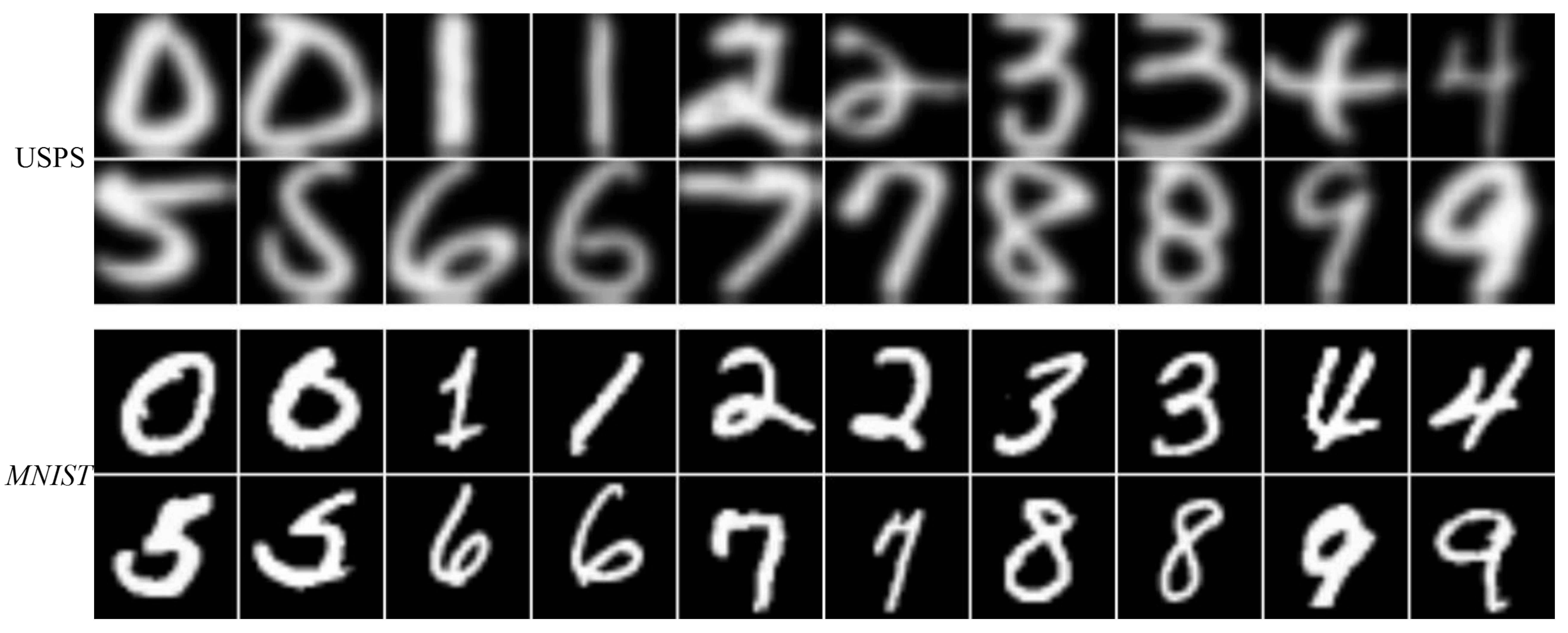}
\label{USPS and MNIST samples}}
\subfigure[\emph{Amazon} and \emph{Webcam} samples]{\includegraphics[ width=0.45\textwidth]{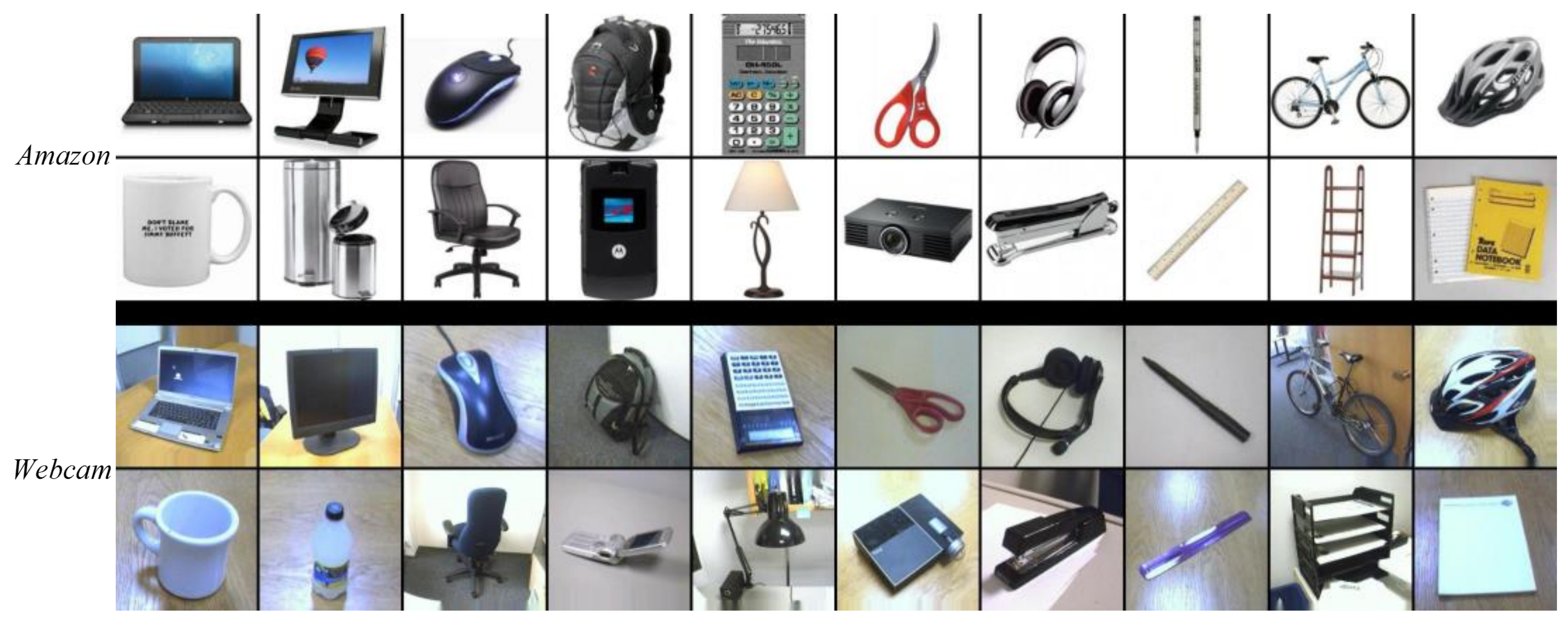}\label{Amazon and Webcam  samples}}
\subfigure[\emph{Pascal} and \emph{Imagenet} samples]{\includegraphics[width=0.45\textwidth]{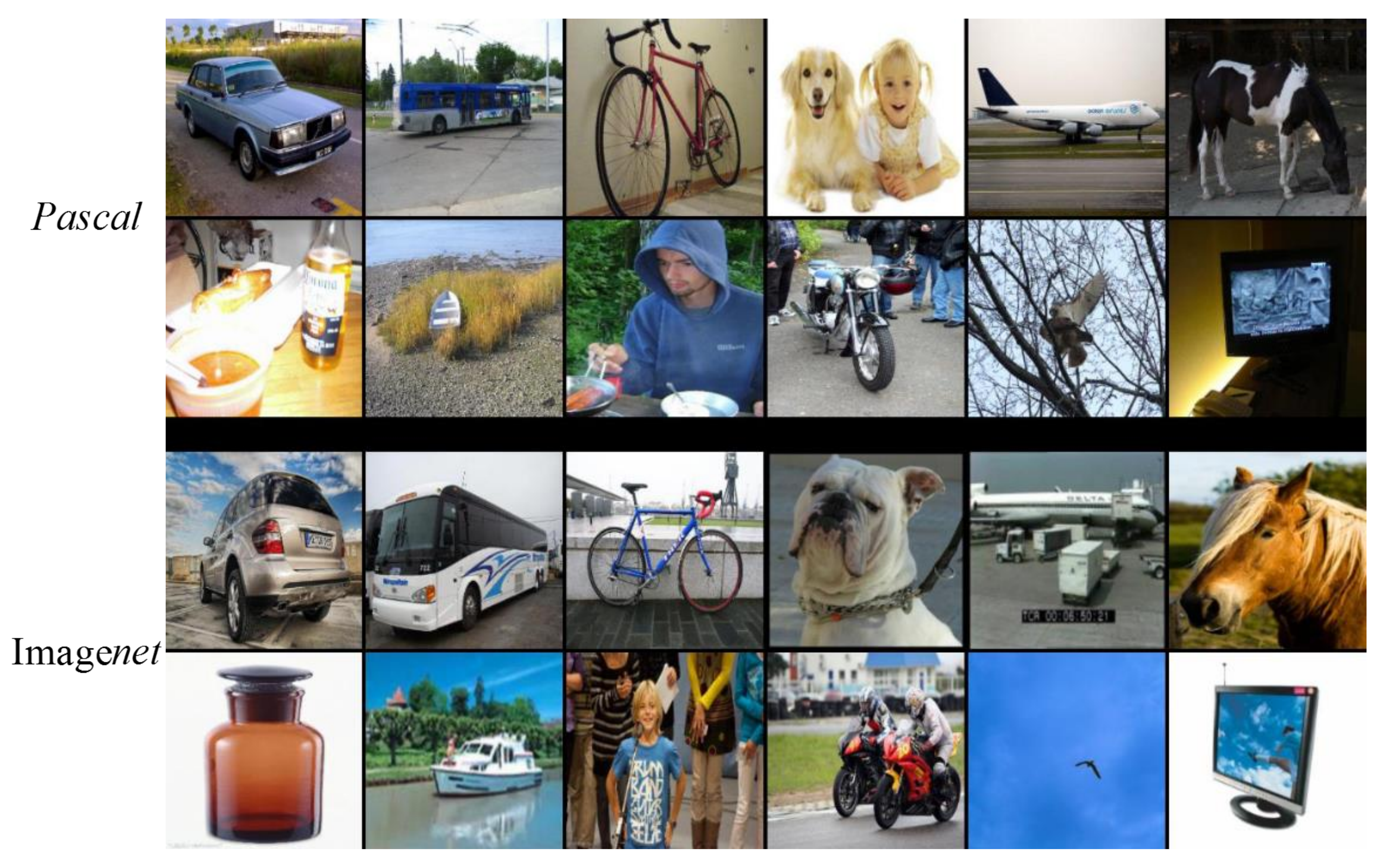}\label{Pascal and Imagenet  samples}}
\caption{Experiment datasets: The first two rows in (a) (b) (c) represent examples from the USPS, \emph{Amazon} and \emph{Pascal} datasets, and the last two rows represent examples from the MNIST,\emph{Webcam} and  \emph{Imagenet} datasets
respectively.}
\vspace{-0.3cm}
\label{figMClass}
\end{figure}

\subsection{Experiment Setup}
Our method is implemented based on Tensorflow.
We use the stochastic gradient descent (SGD) optimizer and set the learning rate  $\eta= \eta_0*\gamma^{[\frac{p}{3000}]}$, where $p$ is the training step varying from 0 to 30000, set $\gamma$  to 0.92, and set $\eta_0 \in \{0.005, 0.01, 0.02, 0.03\}$ for all transfer tasks. We fixed the trade-off regularizer weight $\alpha=0.6$ and domain adaptation regularizer weight $\beta =1.0$ in all experiments.
In order to suppress noisy signal from the domain classifier at early stages during training, we change the hyper-parameter $\lambda$ of domain classification using the following schedule:
$\lambda = \lambda_0*(\frac{2}{1+\exp(-\gamma*q)}-1)$, where $q$ changes from 0 to 1 during progress, and $\lambda_0, \gamma$ are sensitive to different datasets, as discussed in the following paragraphs. 

Specifically, for experiments on \emph{Office-31} and \emph{ImageCLEF-DA} datasets, due to the limited data in the source domain, we fine-tune our model using the pre-trained model of Resnet\footnote{https://github.com/tensorflow/models/tree/master/research/slim} (50 layers) on the Imagenet~\citep{Russakovsky2015ImageNetLS} dataset. Following the notation in ResNet, we fix the convolutional layers of conv1, conv2\_x, and conv3\_x, and fine-tune the rest conv4\_x and conv5\_x. 
Then we train the logits layer of ResNet, the label classifiers and the domain classifier from scratch with learning rate 10 times larger than the fine-tuning part. We set $\gamma = 10$, 
and $\lambda_0 \in \{1.3, 1.5, 1.8, 2.0\}$  for different tasks in \emph{Office-31} while $\lambda_0=1.0$ as a fixed value for tasks in \emph{ImageCLEF-DA}. This progressive strategy significantly stabilizes parameter sensitivity and eases model selection for DART.

For USPS and MNIST datasets, we replace the Resnet with several CNN layers without fine-tuning. We set $\lambda_0=1.0$ , and $\gamma=2.5$. 
Note that for all transfer tasks, we run our method for three times and report the average classification accuracy and the standard error for comparisons.



\subsection{Results}
For MNIST and USPS datasets, the classification results are shown in Table~\ref{USPS MNIST transfer}. The reported results of CoGAN, pixelDA, and UNIT are from their corresponding papers~\citep{Liu2016CoupledGA,Bousmalis2016UPD,Liu2017UnsupervisedIT}. As can be easily observed, the proposed DART  outperforms all baselines
on both tasks. Especially on USPS to MNIST, our model improves the accuracy by a large margin, i.e., about $6\%$, indicating the effectiveness of DART.

\begin{table}[!h]
	\centering
	\caption{Classification accuracy ($\%$) on USPS and MNIST datasets. 
	}
	\begin{tabular}{l|c|c}
            \hline Model& MNIST to USPS & USPS to MNIST\\
            \hline
			CoGAN&95.65&93.15\\\hline
			pixelDA&95.9&-\\\hline
			UNIT&95.97&93.58\\\hline
            \textbf{DART}&$\textbf{98.20}$&$\textbf{99.40}$\\\hline
	\end{tabular}
	\label{USPS MNIST transfer}
\end{table}
\begin{table*}[!t]
    \centering
    \caption{ Classification accuracy ($\%$) on \emph{Office-31} dataset for unsupervised domain adaptation. DART-c denotes the DART network without joint distribution alignment, and DART-s denotes the DART network without label classifier perturbation. Results of the competitive methods are copied from the original literature.}
	\begin{tabular}{l|c|c|c|c|c|c|c}
            \hline Method& $\textbf{A}\to\textbf{W}$ & $\textbf{D}\to\textbf{W}$ & $\textbf{W}\to\textbf{D}$&
            $\textbf{A}\to\textbf{D}$&$\textbf{D}\to\textbf{A}$&
            $\textbf{W}\to\textbf{A}$& Avg\\\hline
            Resnet&$68.4\pm0.2$&$96.7\pm0.1$ &$99.3\pm0.1$ &$68.9\pm0.2$ &$62.5\pm0.3$ &$60.7\pm0.3$ &76.1  \\\hline
            TCA&$72.7\pm0.0$&$96.7\pm0.0$ &$99.6\pm0.0$ &$74.1\pm0.0$ &$61.7\pm0.0$ &$60.9\pm0.0$ &77.6  \\\hline
            GFK&$72.8\pm0.0$&$95.0\pm0.0$ &$98.2\pm0.0$ &$74.5\pm0.0$ &$63.4\pm0.0$ &$61.0\pm0.0$ &77.5  \\\hline
            DDC&$75.6\pm0.2$&$96.0\pm0.2$ &$98.2\pm0.1$ &$76.5\pm0.3$ &$62.4\pm0.4$ &$61.5\pm0.5$ &78.3  \\\hline
            DAN&$80.5\pm0.4$&$97.1\pm0.1$ &$99.6\pm0.1$ &$78.6\pm0.2$ &$63.6\pm0.3$ &$62.8\pm0.2$ &80.4  \\\hline
            RTN&$84.5\pm0.2$ & $96.8\pm0.1$&$99.4\pm0.1$ &$77.5\pm0.3$ &$66.2\pm0.2$ &$64.8\pm0.3$ &81.6  \\\hline
            DANN&$82.0\pm0.4$ &$96.9\pm0.2$ &$99.1\pm0.1$ &$79.7\pm0.4$ &$68.2\pm0.4$ &$67.4\pm0.5$ &82.2  \\\hline
            JAN&$85.4\pm0.3$&$97.4\pm0.2$ &$99.8\pm0.2$ &$84.7\pm0.3$ &$68.6\pm0.3$ &$70.0\pm0.4$ &84.3  \\\hline
            JAN-A&$86.0\pm0.4$ &$96.7\pm0.3$ &$99.7\pm0.1$ &$85.1\pm0.4$ &$69.2\pm0.2$ &$\textbf{70.7}\pm \textbf{0.5}$ &84.6  \\\hline
            DART-c 
            &$84.5\pm0.2$ &$95.6\pm0.1$ &$98.4\pm0.1$ &$82.1\pm0.2$ &$42.4\pm0.2$ &$50.5\pm0.3 $
            & 75.5\\\hline
            DART-s 
            &$85.3\pm0.2$ &$97.6\pm0.1$ &$99.9\pm0.1$ &$86.0\pm0.1$ &$46.7\pm0.2$ &$54.3\pm0.3$
            &78.3\\\hline
            \textbf{DART} &$\textbf{87.3}\pm \textbf{0.1}$ &$\textbf{98.4}\pm \textbf{0.1}$ &$\textbf{99.9}\pm \textbf{0.1}$
            & $\textbf{91.6}\pm \textbf{0.1}$  &$\textbf{70.3}\pm \textbf{0.1}$ &$69.7\pm 0.1$ &$\textbf{86.2}$  \\\hline
	\end{tabular}
	\label{tab:OFFICE_31}
\end{table*}

\begin{table*}[!t]
	\centering
	\caption{Classification accuracy ($\%$) on \emph{ImageCLEF} dataset for unsupervised domain adaptation. Results of the competitive methods are copied from the original literature. }
	\begin{tabular}{l|c|c|c|c|c|c|c}
            \hline Method& $\textbf{I}\to\textbf{P}$ & $\textbf{P}\to\textbf{I}$ & $\textbf{I}\to\textbf{C}$&
            $\textbf{C}\to\textbf{I}$&$\textbf{C}\to\textbf{P}$&
            $\textbf{P}\to\textbf{C}$& Avg\\\hline
            Resnet&$74.8\pm0.3$ &$83.9\pm0.1$ &$91.5\pm0.3$ &$78.0\pm0.2$ &$65.5\pm0.2$ &$91.2\pm0.3$ &80.7  \\\hline
            DAN&$74.5\pm0.4$ &$82.2\pm0.2$ &$92.8\pm0.2$ &$86.3\pm0.4$ &$69.2\pm0.4$ &$89.8\pm0.4$ &82.5  \\\hline
            RTN&$74.6\pm0.3$ &$85.8\pm0.1$ &$94.3\pm0.1$ &$85.9\pm0.3$ &$71.7\pm0.3$ &$91.2\pm0.4$ &83.9  \\\hline
            JAN&$76.8\pm0.4$ &$88.0\pm0.2$ &$94.7\pm0.2$ &$89.5\pm0.3$ &$74.2\pm0.3$ &$91.7\pm0.3$ &85.8  \\\hline
            \textbf{DART}&$\textbf{78.3}\pm \textbf{0.1}$ &$\textbf{89.3}\pm \textbf{0.1}$ &$\textbf{95.3}\pm \textbf{0.1}$ &$\textbf{91.0}\pm \textbf{0.1}$ &$\textbf{75.2}\pm \textbf{0.1}$ &$\textbf{93.5}\pm \textbf{0.1}$ &$\textbf{87.1}$  \\\hline
	\end{tabular}
	\label{tab:ImageCLEF}
\end{table*}

Similarly, the results on \emph{Office-31} and \emph{ImageCLEF-DA} datasets are shown in Tables \ref{tab:OFFICE_31}-\ref{tab:ImageCLEF}. We report the results of Resnet, TCA, GFK, DDC, DAN, RTN, DANN and JAN from~\citep{Long2017DeepTL}, in which all the above algorithms are re-implemented.
For the results of \emph{Office-31} dataset, the proposed DART exceeds the state of the art results around $1.6\%$ in average accuracy. In particular, on $\textbf{A} \to \textbf{D}$, DART achieves a large margin of improvement, i.e., more than $6.0\%$ over the best baseline.


In terms of \emph{ImageCLEF-DA}, the results in Table \ref{tab:ImageCLEF} again demonstrate that our model could achieve higher accuracy than the rest baselines on all transfer tasks. Compared with JAN, our average performance exceeds $1.3\%$.

The DART model outperforms all previous methods and sets new prediction records on most tasks, indicating its superiority of both effectiveness and robustness. 
DART is different from previous methods, since it adapts the joint distribution of high-level features and labels instead of marginal distributions as those in DAN, RTN and DANN, and learns the perturbation function between the label classifiers. These modifications can be the key to the improvement of the prediction performance.



\begin{figure*}[!t]
\centering
\subfigure[DANN: Source domain: \textbf{A}]{
\includegraphics[width=0.24\textwidth]{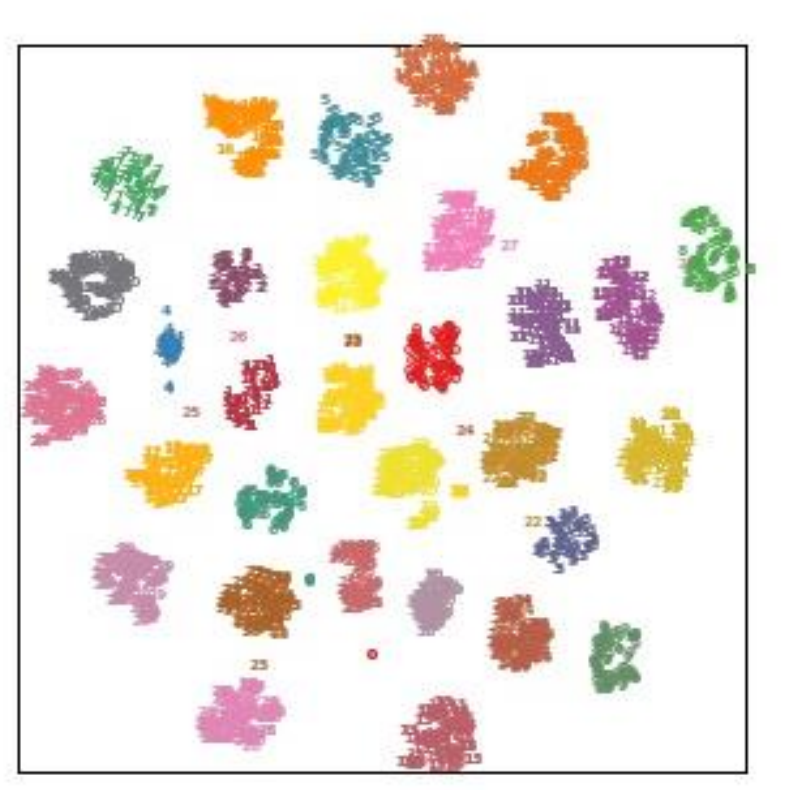}
\label{DANN:A}}
\subfigure[DANN: Target domain: \textbf{W}]{\includegraphics[ width=0.24\textwidth]{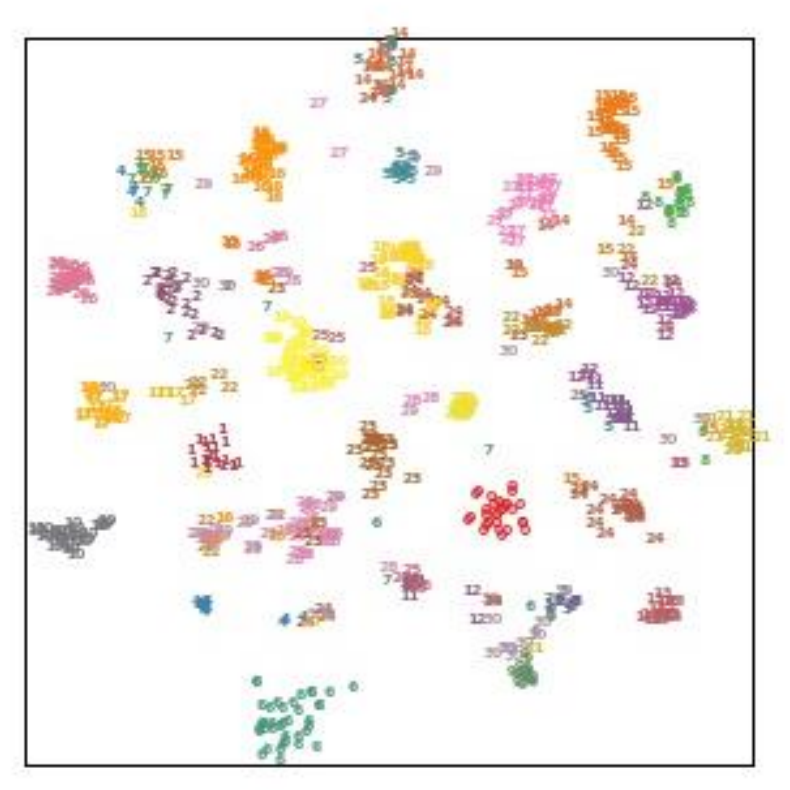}\label{DANN:W}}
\subfigure[DART: Source domain: \textbf{A}]{\includegraphics[ width=0.24\textwidth]{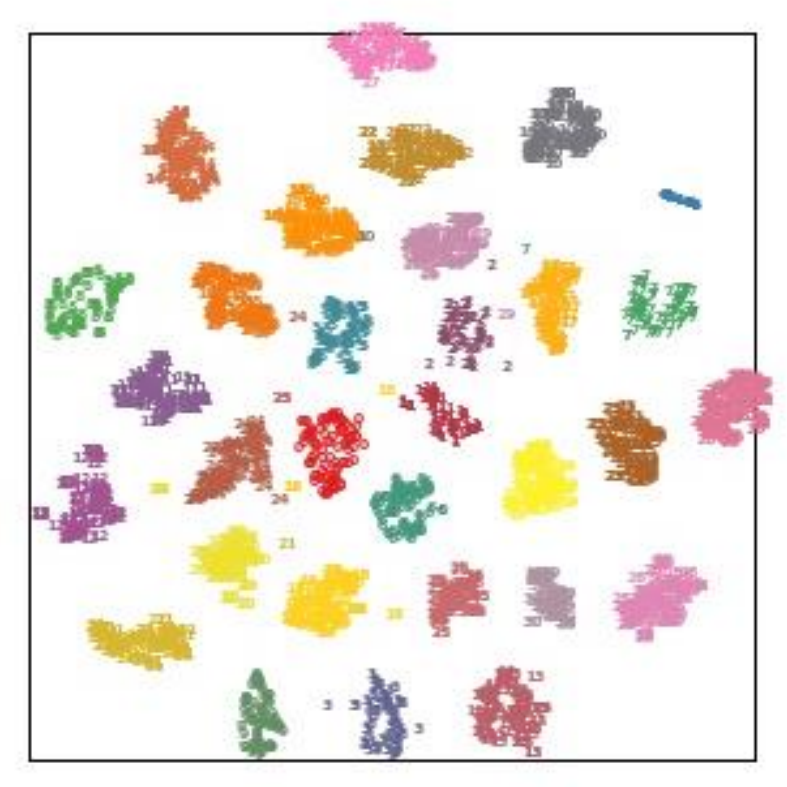}\label{DART:A}}
\subfigure[DART: Target domain \textbf{W}]{\includegraphics[ width=0.24\textwidth]{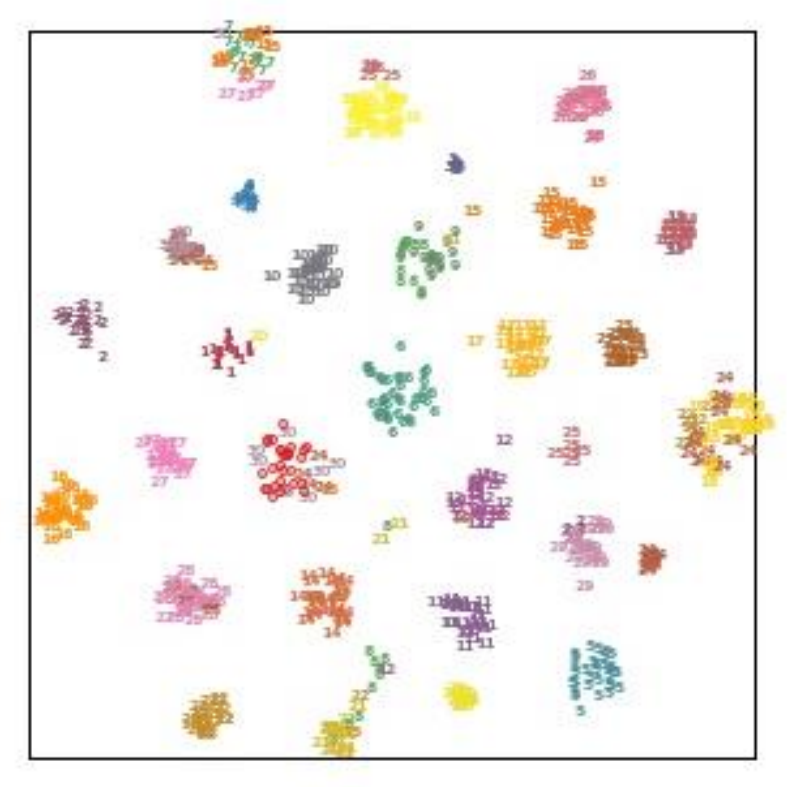}\label{DART:W}}
\caption{The t-SNE visualization of network activations by DANN in (a), (b) and DART in (c), (d) respectively.}
\end{figure*}


\subsection{Results Analysis}
\hspace{0ex}$\textbf{Predictions Visualization:}$
To further visualize our results, we embed the label predictions of DANN and DART using t-SNE~\citep{Donahue2014DeCAFAD} on the example task $\textbf{A}  \rightarrow  \textbf{W}$, and the results are shown in Figure~\ref{DANN:A}-\ref{DART:W} respectively. 
The embeddings of DART show larger margin than those of DANN, indicating better classification performance of the target classifier of DART.
Now it can be observed that the adaptation of joint distribution of features and labels is an effective approach to unsupervised domain adaptation and the modelling of the perturbation between label classifiers is a reasonable extension to previous deep feature adaptation methods. 

\begin{figure*}[!h]
\centering
\subfigure[$\mathcal{A}$-distance]{\includegraphics[width=4.0cm, height= 4.5cm]{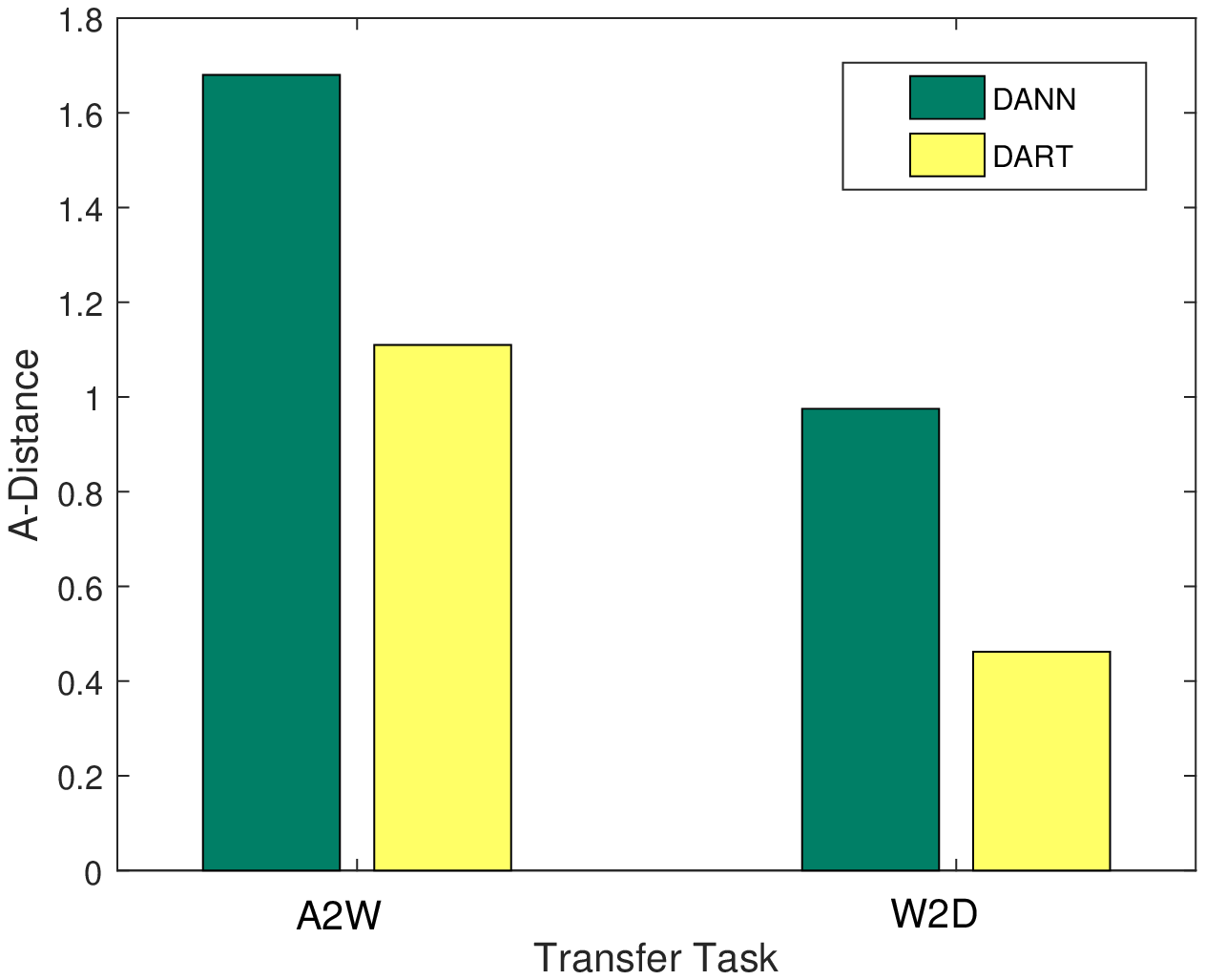}\label{A-distance}}
\subfigure[Convergence]{\includegraphics[width=4.0cm, height= 4.5cm]{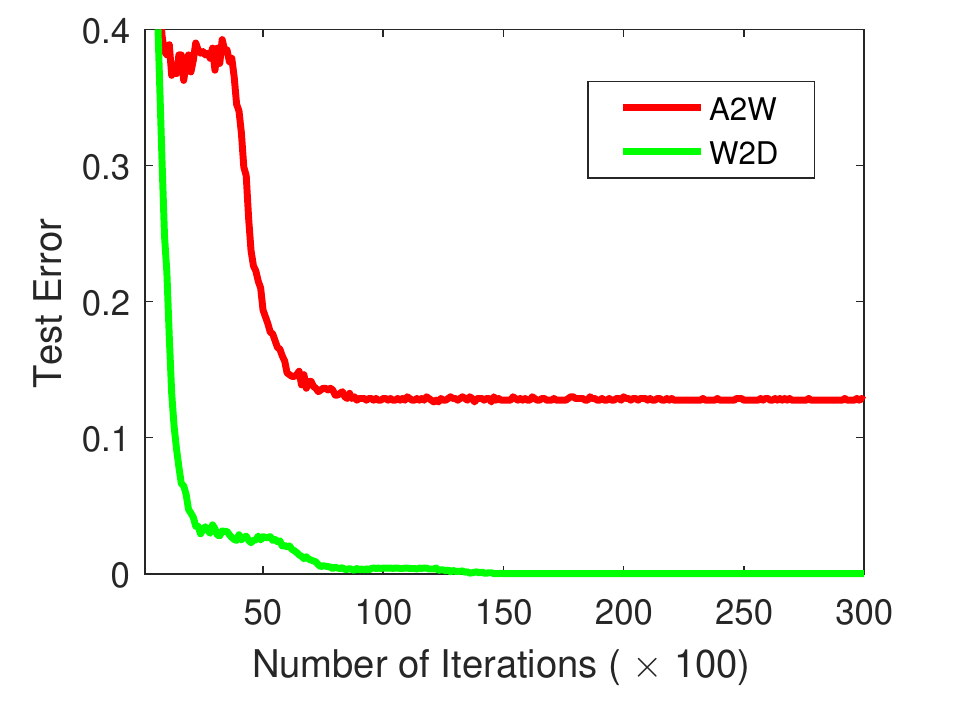}\label{test error}}
\subfigure[Accuracy change on $\alpha$]{ \includegraphics[width=4.0cm, height=4.5cm]{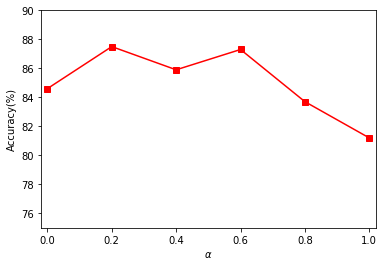}\label{alpha_plot}}
\subfigure[Accuracy change on $\beta$]{\includegraphics[width=4.0cm, height= 4.5cm]{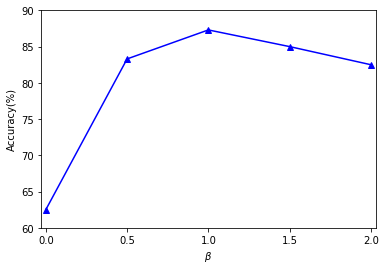}\label{beta_plot}}

\caption{(a) $\mathcal{A}$-distance of DANN and DART; (b) The convergence performance of DART; (c) Accuracy of our model on different choice of $\alpha$, with $\beta$ fixed to 1.0; (d)  Accuracy of the model on different choice of $\beta$, with $\alpha$ fixed to 0.6. }
\label{figMClass}
\end{figure*}

\hspace{-2.8ex}\textbf{Distribution Discrepancy}:
We use $\mathcal{A}$-distance~\citep{BenDavid2009ATO,mansour2009domain} as an alternative measurement to visualize the joint discrepancy attained by DANN and DART.
$\mathcal{A}$-distance is defined as $d_{\mathcal{A}}=2(1-2\epsilon)$, where $\epsilon$ is a generalization error on some binary problems. $\mathcal{A}$-distance could bound the target risk when source risk is limited, i.e., $R_t(G_t)< R_s(G_s)+d_{\mathcal{A}}$,
where $R_t(G_t)=E_{(x,y)\thicksim p(\xt,\yt)}[G_t(G_f(x))\neq y]$ represents the target risk, and $R_s(G_s)=E_{(x,y)\thicksim p(\xs,\ys)}[G_s(G_f(x))\neq y]$ represents the source risk. 
Figure \ref{A-distance} shows $d_{\mathcal{A}}$ values on tasks $\textbf{A}  \rightarrow  \textbf{W}$ and $\textbf{W}  \rightarrow  \textbf{D}$
attained by DANN and DART, respectively. We observe that $d_{\mathcal{A}}$ of DART is much smaller than that of DANN, which suggests that the joint representation of features and labels in DART can bridge different domains more effectively than DANN.

\hspace{-2.8ex}$\textbf{Convergence Performance:}$
To demonstrate the robustness and stability of DART,
we visualize the convergence performance of our model in Figure \ref{test error}
on two example tasks: $\textbf{A} \to \textbf{W}$ and $\textbf{W} \to \textbf{D}$.
Test error of two tasks converges fast in the early 8000 steps and stabilizes in the remaining training process, which testifies the effectiveness and stability of our model.



\hspace{-2.8ex}\textbf{Parameter Sensitivity Analysis:}
We testify the sensitivity of our model to parameters $\alpha$ and $\beta$, i.e. the hyper-parameters for label-classification loss and domain cross-entropy loss.  As an illustration, we use task $\textbf{A}\rightarrow\textbf{W}$ to report the transfer accuracy of DART with different choice of $\alpha$ and $\beta$.  Figure \ref{alpha_plot} reports the accuracy of varying $\alpha \in \{0.0, 0.2, 0.6, 0.8, 1.0\}$ while fixing $\beta = 1.0$. We observe that DART achieves the best performance  when  $\alpha$ is set from [0.2,0.6], although there is a vibration when $\alpha=0.4$. The vibration may be caused by the small number of samples. 
As for $\beta$, the best accuracy can be obtained when setting  $\beta$ from $[0.5, 1.5]$, illustrating a bell-shaped curve as showed in Figure \ref{beta_plot}.


\begin{table*}[h]
	\centering
	\caption{Categories of Unsupervised Domain Adaptation Methods}
	\begin{tabular}{p{6cm}|p{5cm}|p{2cm}|p{2cm}}
    \hline Model & Discrepancy measure & Distribution assumption & Classifier Perturbation\\
    \hline
	  DDC~\citep{Tzeng2014DeepDC} & MMD & Marginal & no\\\hline
	  DAN~\citep{Long2015LearningTF}& MMD & Marginal &no\\\hline
	  DANN~\citep{ganin2016domain}& Domain Adversarial Loss & Marginal & no \\ \hline
      DSN~\citep{Bousmalis2016DomainSN} & MMD/Domain Adversarial Loss & Marginal & no \\\hline
      UNIT~\citep{Liu2017UnsupervisedIT} & GAN & Marginal & no \\\hline
      CoGAN~\citep{Liu2016CoupledGA} & GAN & Marginal & no \\\hline
      DTN~\citep{Taigman2016UnsupervisedCI} & GAN & Marginal & no \\\hline pixelDA~\citep{Bousmalis2016UPD} & GAN & Marginal & no \\\hline
      RTN~\citep{Long2016UnsupervisedDA}  & MMD & Marginal & yes \\\hline
      JDA~\citep{Long2013TransferFL} & MMD & Joint & no \\\hline
      JAN~\citep{Long2017DeepTL} & JMMD & Joint & no \\\hline
      DART(proposed) & Domain Adversarial Loss & Joint & yes \\\hline
	\end{tabular}
	\label{uda:category}
\end{table*}

\section{Related work} \label{sec:related}
Unsupervised domain adaptation based on deep learning architectures can bridge different domains or tasks and mitigate the burden of manual labeling. The goal of domain adaptation is to reduce the domain discrepancy measured in various probability distributions of different domains. In the following, we review different deep domain adaptions methods from three perspectives: (1) domain discrepancy measures; (2) distributions used to measure the discrepancy; (3) differences between label classifiers of both domains. 

In unsupervised domain adaptation, most methods try to learn domain-invariant features, such as DDC~\citep{Tzeng2014DeepDC},
DAN~\citep{Long2015LearningTF}, DANN~\citep{ganin2016domain}, Conditional Adversarial Domain Adaptation~\citep{LongNIPS2018_7436}, etc. That means $p(G_s(\xt)) \approx p(G_t(\xs))$.  Recent approaches assume that both the source domain and the target domain should share a joint distribution of both features and labels. That means the joint distributions of extracted features and labels are shared, i.e,  $p(G_s(\xt),\yt) \approx p(G_t(\xs),\ys)$. Such work include JDA~\citep{Long2013TransferFL} and JAN~\citep{Long2017DeepTL}.
Our work follows the idea of JDA to model the joint distribution, and use the Kronecker project to generate feature and label maps.

In terms of measuring the distribution alignment between the source domain and the target domain, most of previous methods have utilized probabilistic measures, such as the Maximum Mean Discrepancy (MMD) and~\citep{Gretton2006AKM},  the correlation alignment\citep{Sun2016DeepCC}. DDC~\citep{Tzeng2014DeepDC} and DAN\citep{Long2015LearningTF}  minimizes the  discrepancy such that a representation that is both semantically meaningful and domain-invariant can be learned. Recent methods have utilized the idea of adversarial learning to implicitly measure the distribution alignment between the source domain and the target domain.  
Among these methods, UNIT~\citep{Liu2017UnsupervisedIT} and CoGAN~\citep{Liu2016CoupledGA} adopt GANs in their architectures to generate domain translated images and evaluate whether the translated images are realistic for each domain; 
while DTN~\citep{Taigman2016UnsupervisedCI} and pixelDA~\citep{Bousmalis2016UPD} map data in the source domain to the target domain by a generator. DANN~\citep{ganin2016domain} introduces a domain classifier borrowing the idea from
adversarial training to help learn transferable features. The proposed DART follows the idea of DANN to discriminate features from the source domain and the target domain.

In terms of the relation between the source classifier and the target classifier, previous unsupervised domain adaptation methods mostly assume that the same conditional distribution is shared between the target domain and the source domain, i.e.,  $p(\yt| \xt)=p(\ys|\xs)$. Different from this category of approaches, the second category relaxes the rather strong assumption. Instead, it considers a more general scenario in practical applications and assumes that the source classifier and the target classifier differ
by a small perturbation function.
RTN~\citep{Long2016UnsupervisedDA} learns an adaptive classifier by adding residual modules into the source label classifier, fusing the features from multiple layers with a Kronecker product, and then minimizing its discrepancy. 

Taking the advantages of successful unsupervised domain adaptation methods, we design our DART by introducing perturbation to the source classifier and the target classifier to increase flexibility,  measuring the joint distributions of features and labels, and introducing the domain adversarial loss to discriminate two domains.

\section{Conclusion}  \label{sec:conclusion}
This paper presents a novel approach to the unsupervised domain adaptation in deep networks, which enables the end-to-end learning of adaptive classifiers and transferable features. 
Unlike previous methods that match the marginal distributions of features across different domains, the proposed approach reduces the discrepancy of domains using the joint distribution of both high-level features and labels.
In addition, the proposed approach also learns the perturbation function across the label classifiers
via the residual modules, bridging the source classifier and target classifier together to produce more robust outputs.
The approach can be trained by standard back-propagation, which is scalable and can be implemented by most deep learning packages. We conduct extensive experiments on several benchmark datasets, validating the effectiveness and robustness of our model.
Future work constitutes semi-supervised domain adaptation extensions.

\bibliographystyle{plainnat}
\bibliography{egbib}

\end{document}